\documentclass{article}

\usepackage{arxiv}

\usepackage{hyperref}       
\usepackage{url}            
\usepackage{booktabs}       
\usepackage{amsfonts}       
\usepackage{nicefrac}       
\usepackage{microtype}      
\usepackage{lipsum}		
\UseRawInputEncoding
\usepackage{amsmath,amssymb,amsfonts}
\usepackage{subcaption}
\usepackage{graphicx}
\usepackage{float}
\usepackage{todonotes}
\usepackage{cite}
\usepackage{makecell}
\usepackage{tabularx}
\usepackage{multicol}

\DeclareMathOperator{\rel}{rel}

\begin{document}
%
\title{GPR1200: A Benchmark for General-Purpose Content-Based Image Retrieval}
%

%
\author{
   Konstantin Schall, Kai Uwe Barthel, Nico Hezel, Klaus Jung\\
  Visual Computing Group, HTW Berlin\\
  Berlin, Germany \\
  \texttt{konstantin.schall@htw-berlin.de} \\
}

%
%
%
\maketitle              
\begin{abstract}
Even though it has extensively been shown that retrieval specific training of deep neural networks is beneficial for nearest neighbor image search quality, most of these models are trained and tested in the domain of landmarks images. However, some applications use images from various other domains and therefore need a network with good generalization properties - a general-purpose CBIR model. To the best of our knowledge, no testing protocol has so far been introduced to benchmark models with respect to general image retrieval quality. After analyzing popular image retrieval test sets we decided to manually curate GPR1200, an easy to use and accessible but challenging benchmark dataset with a broad range of image categories. This benchmark is subsequently used to evaluate various pretrained models of different architectures on their generalization qualities. We show that large-scale pretraining significantly improves retrieval performance and present experiments on how to further increase these properties by appropriate fine-tuning. With these promising results, we hope to increase interest in the research topic of general-purpose CBIR.
\end{abstract}

\begin{center}
\begin{figure}[ht]
    \makebox[\textwidth]{\includegraphics[width=\textwidth]{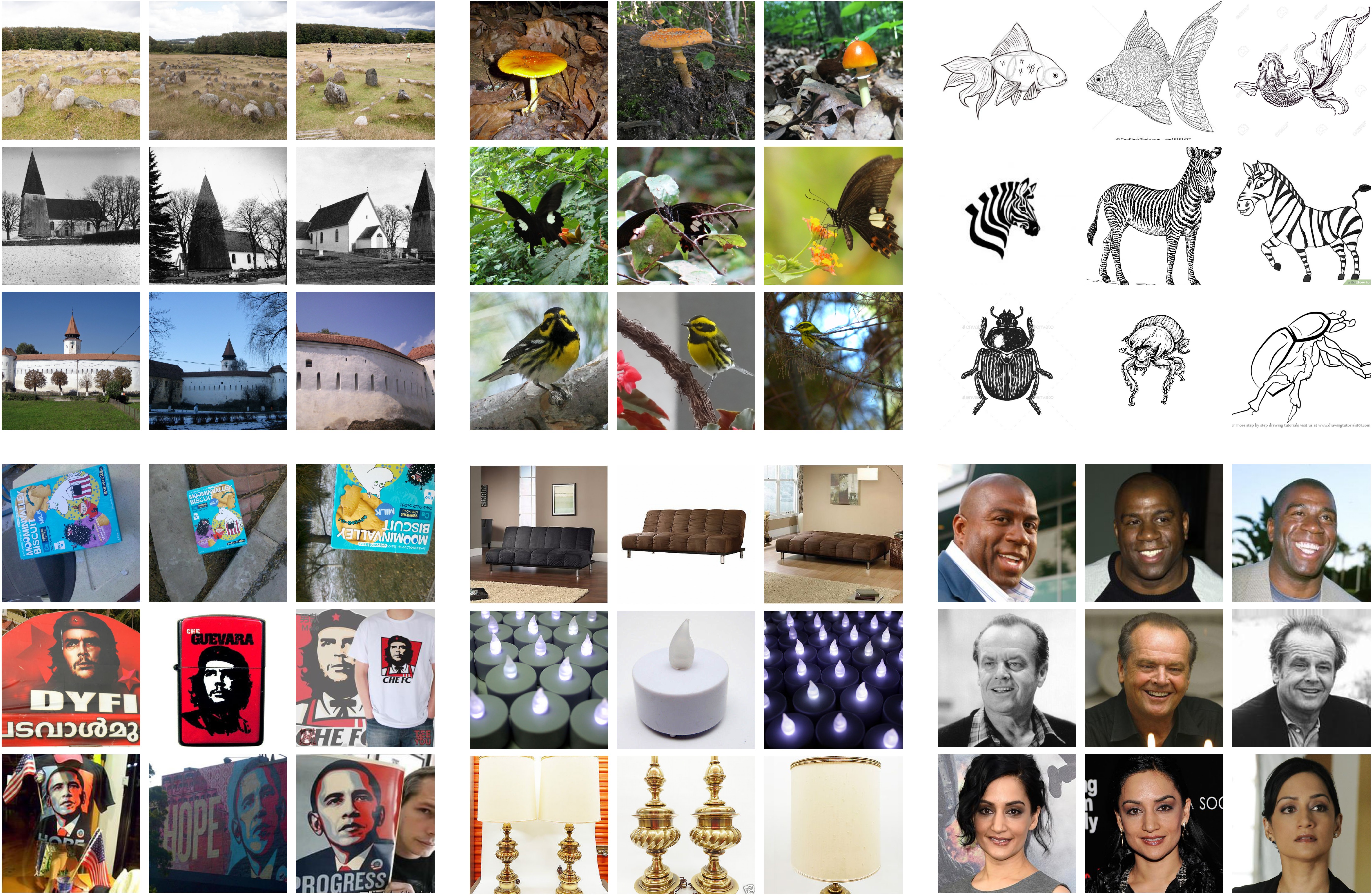}}
    \captionsetup{font=small}
    \caption{Example images from all of the subsets of the newly introduced GPR1200 dataset. Domains from left to right, top to bottom: Landmarks, Nature, Sketches, Objects, Products and Faces} 
    \label{fig:GPR_main}
\end{figure}
\end{center}

\section{Introduction}
Similar to most vision related tasks, deep learning models have taken over in the field of content-based image retrieval (CBIR) over the course of the last decade \cite{conf/cvpr/RadenovicITAC18, conf/eccv/GordoARL16, PCA-W}. More recent research has also shown that performance of retrieval models can further be enhanced by applying a more appropriate loss objective from the metric learning family  which will enforce an embedding space with denser clusters and therefore will often yield better nearest neighbor search results \cite{conf/eccv/GordoARL16, DARAC, DELG}. Other publications introduce more complex pooling functions to obtain even more discriminative image features as a replacement to the commonly used global average pooling \cite{GEM, DARAC}. However, a common denominator of these publications is that the models are trained and tested on domain specific datasets. In the large-scale settings of DELG \cite{DELG} millions of landmark images are used to train a feature extractor model and the feature vectors are evaluated on other landmark datasets of much smaller scale. In the field of deep metric learning the most commonly used datasets are Cars196 \cite{Cars196}, Stanford Online Products \cite{SOP} and CUB200-2011 \cite{CUB_200_2011} which once again are very domain specific. Even though a deep learning model for general-purpose CBIR has a broad range of applications, to the best of our knowledge, no research has been conducted on how to find a suitable network which is not bound to specific domains and generalizes well to all kinds of visual data.  Examples for such applications can be \textit{visual similarity search} for stock photo agencies or \textit{video browsing tools} as used in the Video Browser Showdown \cite{VBS2020}. 

In this paper we try to manifest a protocol for testing deep learning based models for their general-purpose retrieval qualities. After analyzing the currently existing and commonly used evaluation datasets we came to the conclusion that none of the available test sets are suitable for the desired purpose and present the GPR1200 (General Purpose Retrieval) test set. This manually reviewed and well balanced dataset consists of 12000 images with 1200 classes from six different image domains. The goal of this test set is to become an easy to use and accessible benchmark for the research field of general-purpose CBIR. Evaluation code and download instructions can be found at \mbox{\url{http://visual-computing.com/project/GPR1200}}. Furthermore we conducted extensive experiments with a variety of pretrained models of different architectures and show how retrieval qualities can further be increased by appropriate fine-tuning with openly available data. We hope that these results will spark interest in this field and will lead to a healthy competition among researchers which will result in CBIR models with high generalization power in the future.

\section{Related Work}

\paragraph{}

\textbf{CNNs for CBIR.}  Since the introduction of deep learning models for feature extraction in \cite{NeuralCodes} they became the preferred modules in image retrieval systems. Performance of retrieval networks can be evaluated with metrics as \textit{Mean Average Precision} or \textit{Recall}. Both of these metrics will result in a perfect score, if and only if all of the positive images are retrieved and ranked before the negative ones.
Retrieval qualities could further be increased by training neural networks with large scale datasets such as Google Landmarks v2 \cite{GLv2} and the introduction of more sophisticated pooling methods. Instead of the commonly used global average pooling of the activation maps from the last convolutional layer, approaches that aggregate regional information obtained from the activation map into a single, global descriptor have been introduced in \cite{RMAC, DARAC}. Furthermore a very simple but effective pooling method has been shown to produce even better results in \cite{GEM}. The generalized mean pooling (GeM) is practically parameter free and the computational costs are negligible. Additionally, multi-scale feature extraction schemes, where a given image is resized to multiple different resolutions have proven to enhance performance \cite{GEM, DELG, DARAC}. Usually a feature vector is computed for each scale and these vectors are L2-normalized subsequently, summed and L2-normalized again.
However, all of the above approaches only have been applied to images of the landmarks domain and it is unclear if these methods will lead to the same effect in a general-purpose retrieval setting. Another caveat is that these pooling based methods are only known to be useful in the combination with CNN-architectures so far. Since models from the transformer family \cite{VIT, Swin} do not produce conventional activation maps and encapsulate the spatial information of images through self-attention \cite{Attention}, pooling based methods might not be as beneficial here.

\paragraph{}

\textbf{Deep Metric Learning.} The two major streams in this research field can be divided into contrastive and proxy based loss functions. Contrastive loss functions often utilize sampling techniques so that a desired number of examples per class is always available in a batch and then calculate distances of positive feature vectors (from images of the same class) and compare these values to distances of negative ones. The network is harshly penalized if distances between negative feature vectors are smaller than distances between positives \cite{SOP}. Proxy based loss objectives are more similar to the “traditional” Softmax Cross Entropy Loss in the sense that a linear layer is used to map feature vectors to a space where the number of dimensions equals the number of classes in the train dataset. However, the weights \(W_{d,c}\) of these linear layers can be seen as class representation vectors (proxies), one for each class \(c\). The mapping of the \(d\)-dimensional feature vectors is a computation of \(c\) dot products. Most of the loss functions from this family furthermore L2-normalize both the weight matrix \(W_{d,c}\) and the feature vectors, which transforms the linear mapping to a computation of \(c\) cosine-similarities. The goal is to maximize the similarity between feature vectors and its respective class proxy vector and to minimize the similarity to proxy vectors of other classes \cite{ProxyAnchor, ArcFace}.

\section{Analysis of available Datasets}

The number of possible image combinations is almost infinite and it is therefore impossible to perfectly categorize large, web-scale image collections. However, two main categories can be observed while browsing image collections of popular stock photo agencies: photographs and planar images. While the two categories are not mutually exclusive, the major part of available images falls into one or the other. Photographs mostly consist of pictures of real life scenes showing objects such as people, animals, landmarks. Planar images are artificially created and show artworks, sketches, illustrations, book- and web pages, maps or blueprints. At the category level, the most frequently pictured objects were animals, natural and architectural landmarks, people and daily products such as clothes, electrical devices, cars, etc. All of these items are not bound to one of the main categories, since cats, for example, can also be drawn and photographed in real life. Furthermore, all of these categories can also be depicted in an isolated image with a solid color background. An evaluation benchmark dataset for general-purpose CBIR should include a large variety of images and preferably contain examples from both of the observed main categories.  Furthermore, such an image collection should include several of the listed object domains.

\paragraph{}

Another important property of benchmark datasets can be summarized as unique solvability. All images of a given class should contain visual attributes that clearly distinguish them from similar images of other classes. A human annotator should be able to perfectly divide images of all classes and therefore achieve an optimal metric score without being a domain specific expert. Furthermore, the images ideally should not show several objects belonging to different classes nor should the categories have a hierarchical structure. However, as long as these two rules are respected, classes may have different intra-class variances so that a model can be tested on the ability to identify objects on instance and class level. 

\begin{table*}\footnotesize
\begin{center}[ht]
\caption{Comparison of possible CBIR retrieval benchmark datasets}
\begin{tabularx}{\textwidth} { 
   >{\centering\arraybackslash}l 
   >{\centering\arraybackslash}X 
   >{\centering\arraybackslash}X 
   >{\centering\arraybackslash}X 
   >{\centering\arraybackslash}X 
   >{\centering\arraybackslash}X  }
\hline
    \textbf{Dataset}  & \textbf{Domain Variety} & \textbf{Unique Solvability} & \textbf{\# Classes} & \textbf{ \# Images}  \\
     \hline
    Oxford \cite{Oxford} & low & low & 11 & 5k\\
    Paris \cite{Paris} & low & low & 11 & 6k\\
    ROxford \cite{conf/cvpr/RadenovicITAC18} & low & mid & 11 & 5k\\
    RParis \cite{conf/cvpr/RadenovicITAC18} & low & mid & 11 & 6k\\
    
    Google Landmarks V2 \cite{GLv2} & low & low & 200k & 4.1M\\
    \hline
    Holidays \cite{Holidays} & mid & high & 500 & 1.5k\\
    INSTRE \cite{INSTRE} & high & mid & 250 & 28.5k\\
    SOP \cite{SOP} & mid & mid & 22.6k & 120k \\
    CUB200-2011 \cite{CUB_200_2011} & low & high & 200 & 11.8k\\
    Cars196 \cite{Cars196} & low & mid & 196  & 16.2k\\
    iNat Val\cite{iNat} & mid & low & 5.1k  & 96.8k\\
    ImageNet1k Val\cite{ImageNet} & mid & high & 1k  & 50k\\
    ImageNet Sketch Val \cite{ImageNetSketch} & mid & high & 1k  & 50k\\
    \hline
    VGGFace  \cite{VGGFace} & low & mid & 2.6k  & 2.6M\\
    IMDB Faces  \cite{IMDBFaces} & low & low & 20.3k  & 523k\\
\hline
\end{tabularx}
\label{table:datasets}
\end{center}
\end{table*}

\paragraph{}

Table \ref{table:datasets} analyzes a non exhaustive list of possible datasets for benchmarking general-purpose CBIR quality. The criteria \textit{domain variety} and \textit{unique solvabilty} were evaluated according to our experience and observations. The first part of the table contains the most widely used datasets in the landmarks domain. Oxford \cite{Oxford} and Paris\cite{Paris} contain images of 11 different architectural landmarks from the respective city. Five images per class are used as a query and the remaining images serve as the index set. These sets have been revisited in \cite{conf/cvpr/RadenovicITAC18} and it has been shown that the original sets suffered from wrong annotations and suboptimal query selection. Regardless of the version of the datasets, these collections are not suitable to test generalization and general-purpose retrieval, since the number of classes and the domain variety is too low. Google Landmarks v2 \cite{GLv2} is of immense scale but has a very low solvability and domain variety, as a single landmark category can include indoor and outdoor images of the same landmark, as well as examples where the actual landmark is not visible at all. Datasets with more generic image categories are listed in the second section of the table. Even though results for the INRIA Holidays dataset \cite{Holidays} have been reported in the large-scale landmark retrieval field \cite{conf/eccv/GordoARL16, DARAC}, it´s popularity has declined in recent years. The INSTRE \cite{INSTRE} set has the highest of the listed domain varieties, as it consists of categories such as landmarks, illustrations and logos (e.g. corporate and sports). It also includes a large number of photos showing objects (toys, books, etc.) with different backgrounds and degrees of rotation. However, the main downside of this benchmark set is that most of the images do not reflect real-world scenarios and all categories have low intra-class variance. Publications in the metric learning field often use the three datasets CUB200-2011 \cite{CUB_200_2011} (bird species), Cars196 \cite{Cars196} (car models) and Stanford Online Products (SOP) \cite{SOP} (images of products collected from online marketplaces). While the first two datasets have a very low domain variety, SOP is a more fitting candidate with a high number of diverse classes, although it lacks real world images of animals, people and landmarks. iNat provides a very diverse dataset and includes categories from animals, plants, fungi and insects. VGGFaces \cite{VGGFace} is a pure faces dataset with low domain variety. We also explored the idea of using the ImageNet ILSVRC2012 \cite{ImageNet} validation split as a retrieval benchmark, but did not consider this suitable due to the lack of some of the desired domains (landmarks, illustrations, etc.). ImageNet Sketch \cite{ImageNetSketch} contains a collection of sketches and drawings of the same 1000 categories as in ILSVRC2012 but no photographed objects.

\section{General-Purpose Retrieval Benchmark}

The previous analysis has shown that to this day there is no suitable benchmark dataset for general-purpose CBIR. One strategy could be to introduce a protocol that involves testing with several of the listed image collections. Such a procedure would have the disadvantage that the data has to be obtained from multiple sources and that many of the datasets are not uniquely solvable and therefore have different degrees of hardness and cleanness. For these reasons we decided to create a new dataset by manually selecting images from some of the listed collections and classes. We determined that a combination of the following datasets will cover a large part of the domains found in general image collections: 

\begin{enumerate}
    \item Google Landmarks V2 (natural and architectural landmarks)
    \item ImageNet Sketch (black and white sketches of animals and other objects)
     \item iNat (plants, animals, insects and fungi)
    \item INSTRE (planar images and photographs of logos and toys)
    \item SOP (products and objects, partly isolated)
    \item IMDB Faces (human faces)
\end{enumerate}

In order to guarantee a uniform class distribution and solvability of the retrieval tasks, for each subset we decided to manually select 200 classes with 10 examples. We also took care to exclude overlapping classes in this process. Figure \ref{fig:GPR_main} shows some example images of the final dataset.

\textbf{INSTRE.} Only the first 200 classes are included since the last 50 classes show two objects from the other classes in combination. For each of the 200 classes we manually choose ten representatives with different levels of difficulty. 

\textbf{Google Landmarks V2.} Only images from the index set were considered, since those are not meant to be used for training. In addition we discarded all classes with less than twenty examples. This restriction leads to 7k possible classes. Since the original data collection process did not include any manual verification and allowed to have indoor and outdoor scenes under one category, most of the classes did not show ten clearly related examples. For this reason we extracted global feature vectors with the model presented in \cite{GLv2} and further excluded all categories that have less than twenty images which show a minimal cosine similarity of 0.5 among themselves. From the remaining categories the 200 classes and corresponding ten images were manually selected.

\textbf{iNat.} Classes were randomly sampled and images were manually selected according to the described criteria. We excluded some categories from plants and fungi, because they were hard to distinguish by visual characteristics without expert knowledge.

\begin{minipage}{\textwidth}
\textbf{ImageNet Sketch.} Images are only chosen from the validation set in this case. We observed that some of the classes only contain a very small number of unique images which were then mirrored and rotated to fill up the 50 images per class. In the manual selection process of this dataset we tried to avoid these cases and only included a maximum of two variations of the same image. 
\end{minipage}

\textbf{SOP.} We restricted the possible categories to those with at least ten examples and chose images manually according to the specified criteria.

\textbf{IMDB Faces.} The faces part of the original dataset was used for image selection. The celebrities were randomly selected and image instances were manually chosen.

\subsection{GPR1200 Evaluation Protocol}

The resulting collection will be called  \textit{GPR1200} (General-Purpose image Retrieval set with 1200 classes). It represents a compact but complex and challenging benchmarking dataset. As previously stated, the goal is to get a better understanding on how to obtain general-purpose retrieval models. 
As for the evaluation protocol, there is no separation into query and index splits. All of the 12k images are used as queries for robustness. Additionally, no bounding boxes or similar information is given and is not allowed to be used. The chosen evaluation metric is the full mean average precision score which is computed as follows:
\begin{center}
    \begin{tabularx}{\textwidth}{>{\centering\arraybackslash}Xp{0.1cm}>{\centering\arraybackslash}X}
        \begin{equation}
            \label{eq:AP}
            AP(q) = \frac{1}{m_q} \sum_{k=1}^{N} P_q(k)\rel_q(k)
        \end{equation}
        & &
        \begin{equation}
            \label{eq:mAP}
            mAP = \frac{1}{N} \sum_{n=1}^N AP(n)
        \end{equation}
    \end{tabularx}
\end{center}
\(N\) is the number of total images (12k), \(m_q\) equals to the number of positive images (10), \(P_q(k)\) represents the precision at rank \(k\) for the \(q\)-th query and \(\rel_q(k)\) is a binary indicator function denoting the relevance of prediction \(k\) to \(q\) (1 if \(q\) and image at rank \(k\) have the same class and 0 otherwise). Additionally, the mean average precision can be calculated for each of the subsets to get a greater understanding of the models domain capabilities by averaging the 2k subset specific \(AP\) values. The remaining 10k images simultaneously act as distractors.  
It is also important to note that models tested with this benchmarks should not have been trained with any of the datasets used to create GPR1200. However, as for the Google Landmarks v2, the iNat and ImageNet Sketch parts, the validation splits have been used for data selection, training with the respective train splits is therefore perfectly valid. Descriptor post-processing should not be forbidden but if the method is parametrized as in \cite{PCA-W}, the search for parameters must be performed with other data.

\section{Experiments}
In this section we first present metric values according to the stated evaluation protocol for various pretrained deep learning models and discuss the importance of large-scale domain unspecific pretraining for general-purpose CBIR feature extractors. Furthermore, we conducted experiments on how to train general-purpose neural networks for CBIR. All experiments were performed on a single NVIDIA Tesla V100 GPU.

\subsection{Evaluation of pretrained models}

Table \ref{table:results} shows the retrieval results for the evaluated networks which are further described in Table \ref{table:networks}. Solely L2-normalization was used as post-processing for all of these networks. The two evaluated landmark models GLv2 and DELG, which both have been trained on the \textit{Clean} subset of the Google Landmarks V2 dataset \cite{GLv2, DELG} show expected results: high mAPs for the landmarks subset of GPR1200, but relatively low overall mAPs. One of the main findings of the other experiments is that large-scale pretraining with ImageNet21k improves retrieval qualities for all tested network architectures. This can be seen by comparing the R101 to B-R101 (ResNets), DT-B to ViT-B (visual transformers) and the respective versions of the EfficentNetV2 models since the main differences of these pairs are that ImageNet21k pretraining was used in one and omitted in the other. Even though the stated pairs also show discrepancies in the achieved accuracies on the ILSVRC2012 validation split, 80.1 vs. 82.3 for the ResNets, 83.1 vs. 84.5 for the visual transformers and 85.5 vs. 86.3 for the EfficientNets those differences become much clearer by comparing the mean average precisions achieved on GPR1200 (42.8 vs. 55.6, 51.2 vs. 60.0 and 49.0 vs. 62.4). Additionally, it can be seen that for the most architectures a fine-tuning on ILSVRC2012 (denoted as ++ in Table \ref{table:results}) yields better overall mAPs due to the large improvements in the sketches part of the set. This is not surprising since those two sets have the same classes in different representations (sketches vs. photographs). Even though the large variant of ViT achieves the highest overall GPR1200 mAP the Swin-B model seems to have the highest ratio of performance  and latency. This network is also interesting, due to the fact that it achieves higher overall qualities if the ImageNet1k fine-tuning is omitted.  

\begin{table*}\footnotesize
\begin{center}
\caption{Description of evaluated network architectures}
\begin{tabularx}{\textwidth} {
   >{\centering\arraybackslash}l
   >{\centering\arraybackslash}l 
   >{\centering\arraybackslash}X 
   >{\centering\arraybackslash}X
   >{\centering\arraybackslash}X
   >{\centering\arraybackslash}X }
\hline
    \textbf{Key} & \textbf{Network} & \textbf{Params (M)} & \textbf{Latency (ms)} & \textbf{Dim} & \textbf{ImageSize}\\
     \hline
    GLv2 & AF ResNet101 \cite{GLv2} & 44.6 & 72.1 & 2048 & MS* \\
    DELG & DELG ResNet101 \cite{DELG} & 44.6 & 72.1 & 2048 & MS* \\
    \hline
    R101 & ResNet101 \cite{Resnet} & 44.6 & 2.7 & 2048 & 224 \\
    B-R101 & BiT ResNet101 \cite{BIT} & 44.5 & 7.7 & 2048 & 448 \\
    B-R101x3 & BiT ResNet101x3 \cite{BIT} & 387.9 & 41.9 & 2048 & 448 \\
    EfN-L & EfficientNetV2-L \cite{EfficientNetV2} & 118.5 & 10.9 & 1280 & 480 \\
     \hline
    ViT-B & ViT-B \cite{VIT} & 86.6 & 3.9 & 768 & 224 \\
    ViT-L & ViT-L\cite{VIT} & 304.3 & 12.2 & 1024 & 224 \\
    DT-B & DeiT-B  \cite{DEIT} & 86.6 & 3.9 & 768 & 224 \\
    DT-B+D & Distilled DeiT-B  \cite{DEIT} & 87.3 & 3.9 & 768 & 224 \\
    Swin-B & Swin-B  \cite{Swin} & 87.8 & 4.4 & 1024 & 224 \\
    Swin-L & Swin-L  \cite{Swin} & 196.5 & 7.9 & 1536 & 224 \\
    \hline
\end{tabularx}
\label{table:networks}
\vspace{-0.5em}
\caption* {*Landmark models employ a multi scale scheme, where images are used in its original size and resized to one smaller and one larger scale.}
\end{center}
\end{table*}

\begin{table*}
\begin{center}
\caption{GPR1200 mAPs [\%] for evaluated models}
\begin{tabularx}{\textwidth} { 
   >{\centering\arraybackslash}l 
   >{\centering\arraybackslash}X 
   >{\centering\arraybackslash}X
   >{\centering\arraybackslash}X
   >{\centering\arraybackslash}X
   >{\centering\arraybackslash}X 
   >{\centering\arraybackslash}X
   >{\centering\arraybackslash}X
   >{\centering\arraybackslash}X
   >{\centering\arraybackslash}X }
\hline
    \textbf{Key} 
    & \textbf{Train} 
    & \textbf{IN1k Acc}
    & \textbf{GPR \newline mAP}
    & \textbf{Land. \newline mAP}
    & \textbf{Sketch \newline mAP}
    & \textbf{iNat \newline mAP}
    & \textbf{INST \newline mAP}
    & \textbf{SOP \newline mAP}
    & \textbf{Faces \newline mAP} \\
     \hline
Glv2  & \textbf{$--$} &  \textbf{--} & 49.4 & 97.0 & 19.9 & 21.5 & 50.8 & 88.6 & 18.5 \\
DELG  &  \textbf{$--$} &  \textbf{--} & 49.6 & \textbf{97.3} & 20.0 & 21.4 & 51.5 & 88.6 & 18.6 \\
\hline
R101  & \textbf{$-+$} & 80.1 & 42.8 & 65.4 & 29.6 & 27.4 & 42.6 & 74.9 & 17.8 \\
B-R101  & \textbf{$++$} & 82.3 & 55.6 & 82.2 & 47.1 & 43.0 & 52.9 & 86.0 & 22.3 \\
B-R101  & \textbf{$+-$} & \textbf{--} & 54.9 & 81.1 & 41.2 & 41.9 & 51.8 & 87.0 & 26.7 \\
B-R101x3  & \textbf{$++$} & 84.4 & 57.0 & 83.1 & 52.9 & 40.1 & 55.7 & 87.2 & 22.8 \\
B-R101x3  & \textbf{$+-$} & \textbf{--} & 56.0 & 83.4 & 40.8 & 40.1 & 55.1 & 90.0 & 26.6 \\
EfN-L  & \textbf{$-+$} & 85.5 & 49.0 & 69.4 & 57.3 & 32.7 & 44.9 & 71.5 & 17.8 \\
EfN-L  & \textbf{$++$} & \textbf{86.3} & 62.4 & 85.2 & 65.1 & 47.2 & 65.1 & 87.1 & 24.4 \\
EfN-L  & \textbf{$+-$} &  \textbf{--} & 59.2 & 88.2 & 41.5 & 43.8 & 61.4 & \textbf{93.5} & 26.7 \\
    \hline
ViT-B  & \textbf{$++$} & 84.5 & 60.0 & 83.0 & 64.6 & 43.0 & 58.1 & 87.7 & 24.0 \\
ViT-B  & \textbf{$+-$} & \textbf{--} & 57.4 & 84.4 & 46.9 & 37.7 & 58.7 & 90.0 & 26.6 \\
ViT-L  & \textbf{$++$} & 85.8 & \textbf{63.2} & 84.9 & \textbf{74.8} & 45.0 & 60.4 & 88.8 & 25.3 \\
ViT-L  & \textbf{$+-$} & \textbf{--} & 59.9 & 85.7 & 53.0 & 38.1 & 63.5 & 91.2 & 28.3 \\
DT-B  & \textbf{$-+$} & 83.1 & 51.2 & 74.7 & 44.6 & 33.4 & 53.9 & 81.5 & 19.0 \\
DT-B+D  & \textbf{$-+$} & 83.3 & 55.2 & 79.9 & 48.2 & 38.0 & 58.4 & 86.9 & 19.6 \\
Swin-B  & \textbf{$++$} & 85.3 & 61.4  & 84.0 & 65.7 & 41.7 & 63.5 & 86.6 & 24.5 \\
Swin-B  & \textbf{$+-$} & \textbf{--} & 62.9 & 87.4 & 54.2 & \textbf{45.9} & 68.6 & 91.6 & \textbf{29.1} \\
Swin-L & \textbf{$++$} & \textbf{86.3} & 63.0 & 86.5 & 66.1 & 43.2 & 67.8 & 89.4 & 24.5\\
Swin-L & \textbf{$+-$} & \textbf{--} & 63.0 & 88.5 & 50.7 & 44.6 & \textbf{72.5} & 93.1 & 28.5 \\
    \hline
\end{tabularx}
\vspace{-0.5em}
\caption*{
\textbf{$++$}: Pretraining on ImageNet21k and  subsequent fine-tuning on ImageNet1k, \newline\textbf{$+-$}: Training on ImageNet21k only, \textbf{$-+$}: Training on ImageNet1k only, \newline\textbf{$--$}: Neither ImageNet21k nor ImageNet1k have been used for training.
}
\label{table:results}
\end{center}
\end{table*}

\subsection{Training for general-purpose CBIR}

We trained two of the presented model variants with three different datasets. The \textit{clean} subsection of the GLv2 dataset \cite{GLv2} is used in the first setting (denoted as \textit{L} in Table \ref{table:train}), the full ImageNet21k  \cite{ImageNet} set in the second (\textit{I}) and a combination of these two in the third case (\textit{L+I}). Since the ImageNet part has roughly ten times more images, we oversample the landmarks part in the third setting to achieve a balance of these two datasets during training. We conducted experiments with different optimizers and loss functions and only report results for the best combination, which was a standard SGD optimizer with a starting learning rate of 5e-6 and momentum of 0.9 and the ProxyAnchor loss function \cite{ProxyAnchor}. Additionally, the learning rate was scaled by 0.5 after 30\%, 60\% and 90\% of the 200k update steps and a batch size of 128 was used throughout all experiments. 10\% of the train data was used for validation purposes and we present GPR1200 results for the best validation checkpoints. In both cases model variants without ImageNet1k fine-tuning were used, so the effect of the loss function can be seen in the results for the \textit{I}-setting. Although, exactly the same data is used as for the pretraining, the GPR1200 mAP increases in both cases. However, the best overall qualities are obtained with \textit{L+I} training. The GPR1200 mAP of the baseline model Swin-B could be improved by over 3 points in this case. Unsurprisingly, training with the highest variety of data leads to retrieval models with the highest generalization degrees. Training code and model checkpoints can be found at \url{https://this-url-is-hidden.com}.

\begin{table*}\footnotesize
\begin{center}
\caption{GPR1200 mAPs for improved models}
\begin{tabularx}{\textwidth} { 
   >{\centering\arraybackslash}l 
   >{\centering\arraybackslash}X
   >{\centering\arraybackslash}X
   >{\centering\arraybackslash}X
   >{\centering\arraybackslash}X 
   >{\centering\arraybackslash}X
   >{\centering\arraybackslash}X
   >{\centering\arraybackslash}X
   >{\centering\arraybackslash}X }
\hline
    \textbf{Key} 
    & \textbf{Train} 
    & \textbf{GPR \newline mAP}
    & \textbf{Land. \newline mAP}
    & \textbf{Sketch \newline mAP}
    & \textbf{iNat \newline mAP}
    & \textbf{INST \newline mAP}
    & \textbf{SOP \newline mAP}
    & \textbf{Faces \newline mAP} \\
     \hline
B-R101 & -- & 55.6 & 82.2 & 47.1 & 43.0 & 52.9 & 86.0 & 22.3 \\
B-R101 & \textit{L} & 57.6 & 88.2 & 43.2 & 43.4 & 59.8 & 89.6 & 21.4 \\
B-R101 & \textit{I} & 56.4 & 81.6 & 49.4 & 49.2 & 51.8 & 83.6 & 22.8 \\
B-R101 & \textit{L+I} & 58.1 & 87.8 & 47.4 & 47.6 & 58.6 & 85.6 & 21.6 \\
\hline
Swin-B & -- & 62.9 & 87.4 & 54.2 & 45.9 & 68.6 & 91.6 & 29.1 \\
Swin-B & \textit{L} & 65.0  & 91.3 & 53.8 & 46.0 & \textbf{76.8} & \textbf{93.9} & 28.4 \\
Swin-B & \textit{I} & 64.7  & 86.5 & \textbf{64.9} & \textbf{50.3} & 67.8 & 88.4 & \textbf{30.0} \\
Swin-B & \textit{L+I} & \textbf{66.2}  & \textbf{91.7} & 59.2 & \textbf{50.3} & 74.85 & 91.4 & 29.5 \\
    \hline
\end{tabularx}
\caption*{Train: Pretrained models are finetuned with either GLv2Clean \textit{L}, ImageNet21k \textit{I} or a combination thereof \textit{L+I}. -- shows the baseline. }
\label{table:train}
\end{center}
\end{table*}

\label{TrainCBIR}
\section{Conclusion}

This paper has three main contributions: First, the introduction of a new CBIR benchmark dataset that, unlike the existing ones, focuses on high domain diversity to test retrieval systems for their generalization ability. Images were manually selected to ensure solvability and exclude overlaps between categories of different domains. The GPR1200 dataset is easy to use and presents an evaluation protocol that is robust to errors. Secondly, this benchmark was used to evaluate modern deep learning based architectures which showcased that metrics such as ILSVRC2012 validation accuracy can be misleading in the search for general-purpose CBIR models. Lastly, we presented a retrieval-specific training scheme which utilizes a combination of two large-scale datasets to further improve retrieval qualities of two already powerful networks. However, these experiments should be considered rather preliminary, and it is subject to further research whether retrieval-specific pooling or post-processing can lead to further advances. We hope that retrieval models will therefore not only be optimized in a domain-specific setting and will be able to achieve higher levels of generalization in the future. 

%
%
%
\bibliographystyle{unsrt}
\bibliography{references}

\end{document}